\RequirePackage[T1]{fontenc}
\documentclass[10pt,twocolumn,letterpaper]{article}

\usepackage{wacv}
\usepackage{multirow}
\newcommand{\FinalGupdmAuthors}{%
\scalebox{0.97707852}[1]{P\kern-0.014em an\kern0.160em Mu,\kern0.1705em Hanning\kern0.160em X\kern0em u,\kern0.1705em Zhe\kern0em yuan\kern0.160em Liu,\kern0.171em Zheng\kern0.160em W\kern-0.080em ang,\kern0.171em Sixian\kern0.161em Chan,}\newline
and Cong Bai}

\definecolor{wacvblue}{rgb}{0.21,0.49,0.74}
\usepackage[pagebackref,breaklinks,colorlinks,allcolors=wacvblue]{hyperref}

\newcommand{\netname}{SandBAR}
\newcommand{\spark}{SPARK}
\newcommand{\past}{PaST}
\newcommand{\sparkfull}{Split-band Parallel Asymmetric Repair Kernel}
\newcommand{\pastfull}{Post-attention Scaffold Training}

\def\wacvPaperID{1450}
\def\confName{WACV}
\def\confYear{2027}

\title{\netname{}: A Band-Asymmetric Architecture for Underwater Image
Restoration with Removable Frequency Scaffolding}
\author{%
Yanzi Liu$^{1}$\quad
Haoran Hou$^{2}$\setcounter{footnote}{1}\thanks{Haoran Hou, Jiachi Tong, and Shen Chen contributed equally.}\quad
Jiachi Tong$^{3}$\footnotemark[2]\quad
Shen Chen$^{1}$\footnotemark[2]\quad
Shuang Chen$^{4}$\setcounter{footnote}{0}\thanks{Corresponding author.}\\
$^{1}$School of Biomedical Engineering \& Imaging Sciences, King's College London, London, UK\\
$^{2}$Institute of Psychiatry, Psychology \& Neuroscience, King's College London, London, UK\\
$^{3}$Department of Mathematics, Imperial College London, London, UK\\
$^{4}$Department of Computer Science, Durham University, Durham, UK\\
{\tt\small \{yanzi.liu,haoran.hou,shen.chen\}@kcl.ac.uk\quad jt2522@ic.ac.uk\quad shuang.chen@durham.ac.uk}
}
\date{}

\begin{document}
\maketitle


\newcommand{\OursEUVPPSNR}{29.89}
\newcommand{\OursEUVPSSIM}{0.910}
\newcommand{\PlainEUVPPSNR}{29.50}
\newcommand{\PlainEUVPSSIM}{0.907}
\newcommand{\PaSTEUVPGain}{0.39}
\newcommand{\OursLSUIPSNR}{28.92}
\newcommand{\OursLSUISSIM}{0.922}
\newcommand{\PlainLSUIPSNR}{28.52}
\newcommand{\PlainLSUISSIM}{0.919}
\newcommand{\PaSTLSUIGain}{0.40}
\newcommand{\OursUIEBPSNR}{22.77}
\newcommand{\OursUIEBSSIM}{0.914}
\newcommand{\PlainUIEBPSNR}{22.34}
\newcommand{\PlainUIEBSSIM}{0.905}
\newcommand{\PaSTUIEBGain}{0.43}
\newcommand{\OursMeanPSNR}{27.19}
\newcommand{\PlainMeanPSNR}{26.79}
\newcommand{\PaSTMeanGain}{0.41}
\newcommand{\PublishedMeanMargin}{1.0}


\newcommand{\AblFullPSNR}{26.72}
\newcommand{\AblGsplitPSNR}{25.73}
\newcommand{\AblGsplitDropPct}{3.69}
\newcommand{\AblGsplitSSIMDropPct}{1.02}
\newcommand{\AblSymopsPSNR}{26.29}
\newcommand{\AblSymopsDropPct}{1.63}
\newcommand{\AblSymopsSSIMDropPct}{0.28}
\newcommand{\AblNoLFPSNR}{26.28}
\newcommand{\AblNoLFDropPct}{1.65}
\newcommand{\AblNoLFSSIMDropPct}{0.57}
\newcommand{\AblGuidanceAlone}{+0.02}

\providecommand{\PLACEHOLDERTRAJ}{\textbf{??}}

\begin{abstract}
Reliable underwater imagery supports marine monitoring, ecological assessment
and autonomous exploration, yet light scattering, absorption and turbidity
jointly degrade image clarity, distort colour and suppress fine structure.
Although these effects are physically interwoven, they commonly appear as
complementary frequency tendencies: large-scale, channel-dependent appearance
distortions and local high-frequency detail loss. We refer to this tendency as
band asymmetry; it motivates different forms of restoration for global colour
and local structure. Our primary contribution is \netname{}, a novel unified
band-asymmetric architecture for underwater image restoration. Rather than
treating frequency processing as an isolated module, \netname{} organises the
restoration pipeline around band asymmetry at the block, encoder and objective
levels. Within this architecture, we introduce a novel band-asymmetric sandwich
block with a reserved post-attention repair slot and the \sparkfull{} (\spark{}),
a novel module that applies
channel-aware large-support correction to the low band, which predominantly
carries large-scale appearance distortions, and lightweight local refinement to
the detail-bearing high band, thereby restoring global colour while preserving
fine structure. To deliver the global cast to the depth at which the network
represents global structure, a learnable
low-frequency guidance path injects a low-pass view of the input into the deep
encoder, and a three-scale spatial-frequency objective supervises the same axis
throughout the decoder. Building on the reserved post-attention slot, we
further introduce \pastfull{} (\past{}), a novel removable nonlinear scaffolding
scheme that trains
the permanent \netname{} graph under band-specific capacity and runs it at
inference without that capacity. \netname{} reaches \OursEUVPPSNR{} / \OursLSUIPSNR{} /
\OursUIEBPSNR{}\,dB on EUVP / LSUI / UIEB and establishes a new three-domain
mean state of the art on the benchmark family,
exceeding the strongest published result by approximately
\PublishedMeanMargin{}\,dB. \past{} adds no parameters or FLOPs at inference and
preserves matched latency.
\end{abstract}

\section{Introduction}
\label{sec:intro}

\begin{figure}[t]
\centering
\includegraphics[width=\linewidth]{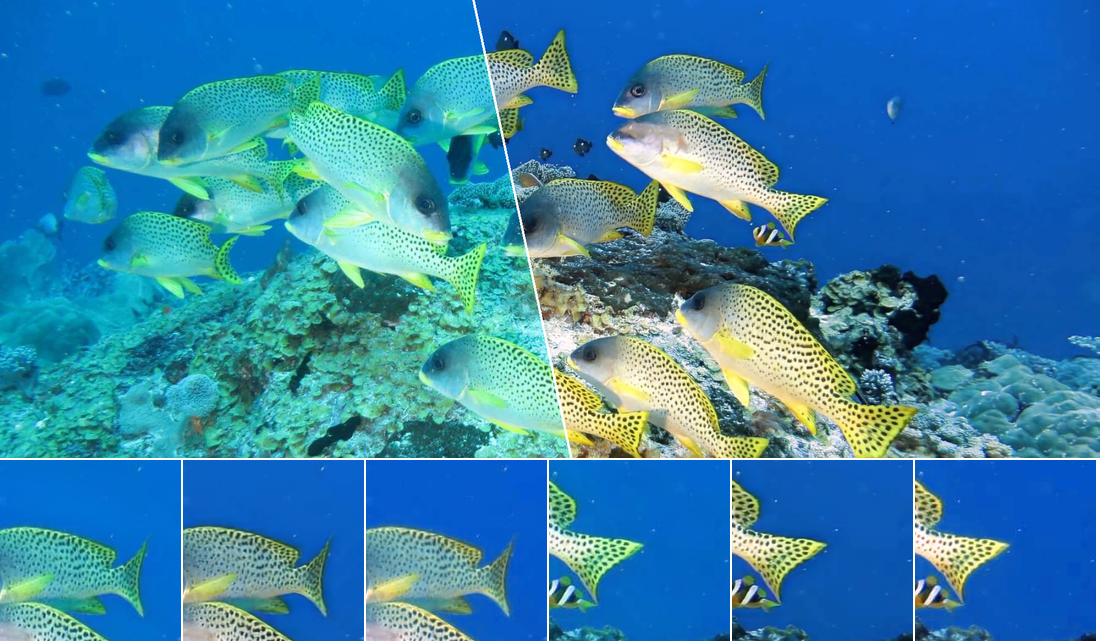}
\par\vspace{1pt}
{\fontsize{5.0}{5.8}\selectfont
\makebox[\dimexpr\linewidth/6\relax][c]{(a) UIR-PolyKernel}%
\makebox[\dimexpr\linewidth/6\relax][c]{(b) \netname{}}%
\makebox[\dimexpr\linewidth/6\relax][c]{(c) Ground Truth}%
\makebox[\dimexpr\linewidth/6\relax][c]{(d) UIR-PolyKernel}%
\makebox[\dimexpr\linewidth/6\relax][c]{(e) \netname{}}%
\makebox[\dimexpr\linewidth/6\relax][c]{(f) Ground Truth}}
\caption{\textbf{Visual comparison of \netname{} and a competing method.}
Top: input and \netname{}. \netname{} restores global colour while preserving
texture detail.}
\label{fig:teaser}
\vspace{-1.0\baselineskip}
\end{figure}

\looseness=-1
Underwater image degradation is not a single uniform loss of visual quality.
Light scattering, absorption and turbidity jointly reduce visibility, distort
colour and blur fine structure. Their effects are physically interwoven:
attenuation and backscatter vary with water conditions, scene range and content,
so no single mechanism belongs exclusively to one spatial-frequency
band~\cite{akkaynak2018revised,akkaynak2019seathru}. In the image domain,
however, the resulting degradation often exhibits complementary frequency
tendencies. Large-scale, channel-dependent distortions dominate background
colour and smooth appearance, whereas edges, textures and small objects
concentrate high-frequency structure and lose contrast first. We refer to this
empirical tendency as \textbf{band asymmetry}. It motivates smooth, channel-structured
correction for the low-frequency component and local detail refinement for the
high-frequency component.
Figure~\ref{fig:teaser} illustrates this behaviour: \netname{} corrects the
large-scale colour cast while preserving fine object texture and boundaries.

\looseness=-1
Deep underwater restoration has advanced rapidly, from physics-guided
CNNs~\cite{li2019underwater,li2021underwater} through
transformers~\cite{peng2023ushape,zamir2022restormer} to recent
spatial--spectral state-space models~\cite{ssuie2025}. Band asymmetry
nevertheless enters these architectures in one of two ways. Uniform backbones
leave it implicit and rely on a homogeneous stack to discover the distinction
during training. Frequency-aware designs~\cite{cui2023sfnet,kong2023fftformer}
model it explicitly through decomposition and specialised per-band processing.
These approaches generally treat band asymmetry either implicitly or within an
individual component, rather than organising block topology, encoder guidance
and supervision around it as one underwater restoration architecture. Moreover,
explicit frequency operations remain active at inference, so their additional
branches and band-specific transformations must be executed for every restored
image. This creates both an architectural gap and a practical tension between
explicit frequency modelling and efficient inference.

These limitations motivate our primary goal: to build a unified underwater
restoration architecture that makes band asymmetry explicit throughout the
restoration pipeline. A further challenge is whether the nonlinear band-specific
capacity used by such an architecture must remain in the network at inference.
Content-adaptive decomposition, asymmetric per-band processing and gated
recombination are nonlinear operations, so they cannot be algebraically folded
into neighbouring layers in the same way as linear re-parameterisation branches
can~\cite{ding2021repvgg,ding2021dbb}.

\looseness=-1
Our main contribution is \netname{}, a novel unified band-asymmetric
architecture for underwater image restoration. It translates this tendency into
three coordinated levels of design. At the block level, a
sandwich transformer block gives every stage two feed-forward slots---one before
attention for generic mixing, one after attention reserved for band-specific
processing---where a standard restoration block~\cite{zamir2022restormer} has
one. At the encoder level, a learnable low-frequency guidance path delivers the
global cast to the depth at which the network represents global structure. At
the objective level, three decoder heads are supervised in both the spatial and
the frequency domain. At its core is the \sparkfull{} (\spark{}), a novel
module that occupies the reserved slot: it applies channel-aware large-support
correction to the low band, which predominantly carries large-scale appearance
distortions, and lightweight local refinement to the detail-bearing high band,
thereby restoring global colour while preserving surviving detail.

\begin{figure}[t]
  \centering
  \includegraphics[width=0.90\linewidth]{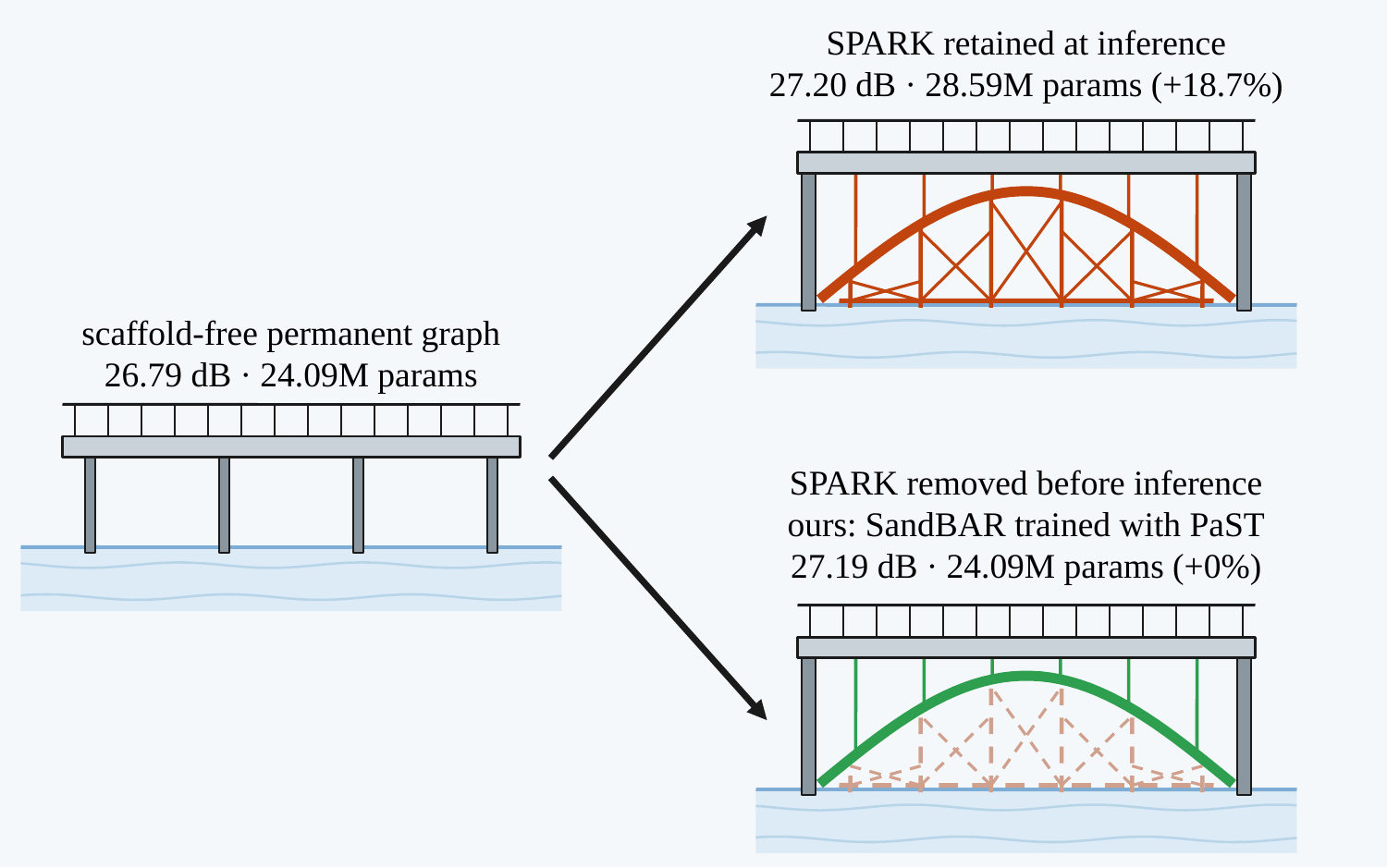}
  \caption{\textbf{\past{} preserves performance while reducing parameters
  and inference cost.} Removing \spark{} before inference restores the
  permanent graph's parameter count, FLOPs and latency.}
  \label{fig:past_tradeoff}
\end{figure}

\looseness=-1
To make this architecture inference-efficient, we further introduce
\pastfull{} (\past{}), a novel removable nonlinear scaffolding scheme within
\netname{}. \spark{} modulates a residual around the
slot input, so removing it collapses the block onto its scaffold-free form and
the single constant it leaves behind folds into the adjacent normalisation.
Figure~\ref{fig:past_tradeoff} summarises the resulting trade-off. Relative to the
separately trained scaffold-free permanent graph, \past{} raises the
three-domain mean PSNR from \PlainMeanPSNR{} to \OursMeanPSNR{}\,dB at the same
$24.09$M-parameter inference size. Retaining \spark{} at inference increases
the graph to $28.59$M parameters and reaches $27.20$\,dB, only
$0.0036$\,dB above the post-fold result using the unrounded values.
\netname{} reaches \OursEUVPPSNR{} / \OursLSUIPSNR{} / \OursUIEBPSNR{}\,dB on
EUVP / LSUI / UIEB and attains a three-domain mean of \OursMeanPSNR{}\,dB,
exceeding the strongest
published result on the benchmark family by approximately
\PublishedMeanMargin{}\,dB. \past{} obtains this with no added parameters or
FLOPs at inference and at matched latency.

\noindent Our contributions are:
\begin{itemize}\itemsep2pt
  \item \textbf{Our primary contribution is \netname{}, a novel unified
  band-asymmetric architecture for underwater image restoration.} It makes band
  asymmetry an explicit architectural prior and instantiates it at the block,
  encoder and objective levels:
  sandwich blocks reserving a post-attention slot for band repair, a learnable
  low-frequency guidance path into the deep encoder, and a three-scale
  spatial-frequency objective. The three levels operate as one system organised
  around a single axis.

  \item \textbf{Within \netname{}, we introduce a novel band-asymmetric
  sandwich block and the novel \spark{} split-band repair module.} The block reserves a
  post-attention slot for band repair, while \spark{} treats the two frequency
  bands with operators matched to their dominant content.

  \item \textbf{We further introduce \past{}, a novel removable nonlinear
  scaffolding scheme for training \netname{}.} Band-specific capacity shapes the
  network during joint training, after which the scaffold is removed and the
  permanent \netname{} graph runs inference without it.

  \item \textbf{\netname{} establishes a new three-domain mean state of the art
  on the benchmark family}, exceeding the strongest published result by
  approximately \PublishedMeanMargin{}\,dB and obtaining the \past{} benefit
  with no added parameters or FLOPs at inference
  and at matched latency.
\end{itemize}

\section{Related work}
\label{sec:related}

\paragraph{Underwater image restoration.}
Early learning-based methods paired physical priors with CNNs:
Water-Net~\cite{li2019underwater} fuses pre-processed inputs under a gated
network, Ucolor~\cite{li2021underwater} embeds a medium-transmission prior into
a multi-colour-space encoder, and FUnIE-GAN~\cite{islam2020fast} targets
real-time enhancement adversarially. Transformer architectures raised accuracy
at higher cost: the U-shape Transformer~\cite{peng2023ushape} introduced the
LSUI benchmark, and general restoration backbones such as
Restormer~\cite{zamir2022restormer} transfer strongly to the task. Recent work
explores semi-supervised learning~\cite{huang2023semiuir} and
spatial--spectral state-space models~\cite{ssuie2025}. Across this line,
underwater-specific capacity generally remains part of the permanent inference
graph. \netname{} instead makes band asymmetry an architecture-wide organising
principle across block topology, encoder guidance and supervision. Within this
architecture, task-specific frequency capacity shapes the network during
training without becoming permanent inference cost.

\paragraph{Frequency-aware restoration.}
Decomposing features by frequency is an established design in general
restoration: SFNet~\cite{cui2023sfnet} selects informative bands with
content-adaptive filters, and frequency-domain transformers~\cite{kong2023fftformer}
compute attention spectrally. \netname{} instantiates its band split with
SFNet's content-adaptive low-pass filter~\cite{cui2023sfnet} and its gated
recombination. Within the overall \netname{} architecture, \spark{} advances
frequency-aware restoration by assigning different operators to the low- and
high-frequency bands according to their dominant content rather than treating
them symmetrically. \past{} further makes the complete in-block frequency path
training-only rather than permanent.

\paragraph{Training-time-only capacity.}
\looseness=-1
Three families discard capacity between training and inference. Structural
re-parameterisation---RepVGG~\cite{ding2021repvgg}, Diverse Branch
Blocks~\cite{ding2021dbb}, edge-oriented blocks for
super-resolution~\cite{zhang2021ecbsr}---trains multi-branch \emph{linear}
topologies that are algebraically collapsed into single convolutions. Deep
supervision~\cite{lee2015deeply} discards auxiliary heads attached
\emph{outside} the backbone. Distillation~\cite{hinton2015distilling} discards
an entire teacher network but requires one to exist. Unlike linear structural
re-parameterisation, external auxiliary heads or teacher--student distillation,
\past{} is the training mechanism within \netname{} that trains a single network
under removable \emph{nonlinear} in-block scaffolding and, after one joint
optimisation run, retains only its permanent graph at inference.

\section{\netname{} architecture}
\label{sec:method}

\begin{figure*}[t]
  \centering
  \includegraphics[width=\linewidth,trim=0bp 3.3bp 0bp 3.5bp,clip]{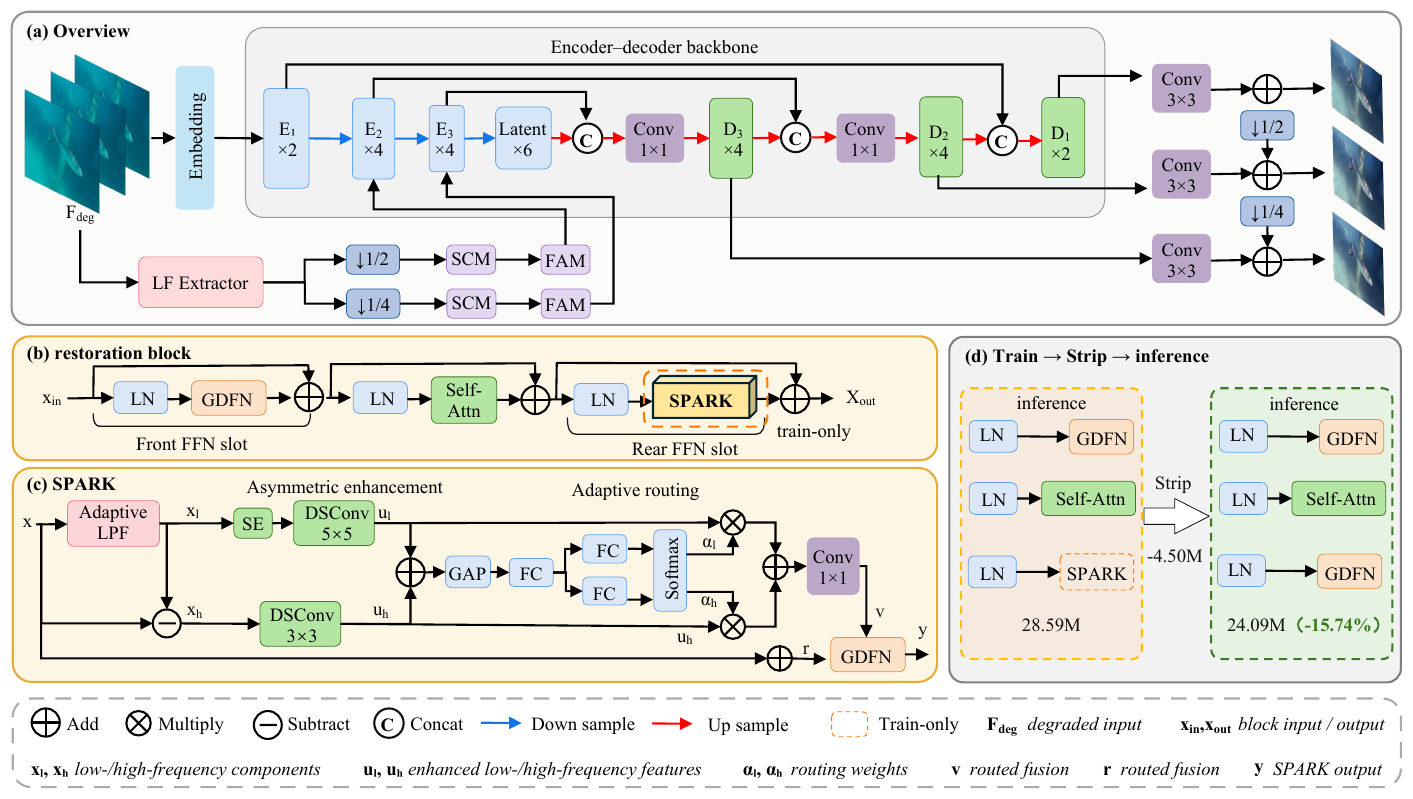}
  \caption{\textbf{\netname{} architecture and its constituent designs.}
  \textbf{(a)} Encoder--decoder
  backbone with a learnable low-frequency guidance path injected at encoder
  levels 2 and 3, and three supervised output heads. \textbf{(b)} The sandwich
  block: two feed-forward slots around attention, with \spark{} occupying only
  the post-attention slot. \textbf{(c)} \spark{}: content-adaptive band split,
  asymmetric per-band repair, gated recombination and residual closing.
  \textbf{(d)} \past{}: after joint training the scaffold is removed and its
  bias folded, and what remains at inference is the permanent scaffold-free
  block.}
  \label{fig:arch}
\end{figure*}

\looseness=-1
Figure~\ref{fig:arch} presents the complete \netname{} architecture and shows
how its constituent designs form one coordinated band-asymmetric restoration
system. \netname{} translates band asymmetry into three levels of design: a
post-attention slot for asymmetric band repair
(Sec.~\ref{sec:method_block}--\ref{sec:method_spark}), a low-frequency guidance
path for global colour correction (Sec.~\ref{sec:method_lf}), and a three-scale
spatial-frequency objective (Sec.~\ref{sec:method_loss}). \spark{} occupies the
reserved slot during training, while \past{} removes its frequency machinery
before inference (Sec.~\ref{sec:method_past}).

\subsection{Band-asymmetric sandwich block}
\label{sec:method_block}

To give band repair a dedicated position after global context mixing, every
\netname{} stage uses a sandwich block with feed-forward slots before and after
attention:
\begin{align}
  \mathbf{x}_1 &= \mathbf{x}_0 + \mathrm{G}_{\mathrm{pre}}
                  (\mathrm{LN}(\mathbf{x}_0)),\label{eq:pre}\\
  \mathbf{x}_2 &= \mathbf{x}_1 + \mathrm{A}(\mathrm{LN}(\mathbf{x}_1)),
                  \label{eq:attn}\\
  \mathbf{x}_3 &= \mathbf{x}_2 + \mathrm{S}(\mathrm{LN}(\mathbf{x}_2)),
                  \label{eq:post}
\end{align}
where $\mathrm{A}$ is multi-Dconv-head transposed
attention~\cite{zamir2022restormer}, $\mathrm{G}_{\mathrm{pre}}$ is a
gated-Dconv FFN (GDFN)~\cite{zamir2022restormer}, and $\mathrm{S}$ is the
post-attention slot. A standard restoration block has a single feed-forward
network; reserving a second slot separates band repair from generic channel
mixing and places it where global context has just been aggregated. In the
inference graph the slot holds a second GDFN; during training it holds \spark{}.
Both slots sit on residual branches, which is what makes the slot's content
removable (Sec.~\ref{sec:method_past}). This sandwich topology forms the
block-level core of \netname{}.

\subsection{\spark{}: asymmetric split-band repair}
\label{sec:method_spark}

To restore global colour without suppressing surviving detail, \spark{} applies
different operators to the low- and high-frequency components of each
post-attention feature. Given the slot input
$\mathbf{z}=\mathrm{LN}(\mathbf{x}_2)$, a content-adaptive low-pass
filter~\cite{cui2023sfnet} predicts a per-image, softmax-normalised
$K{\times}K$ kernel and produces
\begin{equation}
  \mathbf{z}_l = \mathcal{F}_{\mathrm{lp}}(\mathbf{z}), \qquad
  \mathbf{z}_h = \mathbf{z} - \mathbf{z}_l,
  \label{eq:split}
\end{equation}
an exact two-band partition. The bands then receive asymmetric treatment:
\begin{align}
  \mathbf{u}_l &= \mathrm{PW}\!\left(\mathrm{DW}_{5\times5}\!\left(
                  \mathrm{SE}(\mathbf{z}_l)\right)\right), &
  \mathbf{u}_h &= \mathrm{PW}\!\left(\mathrm{DW}_{3\times3}
                  (\mathbf{z}_h)\right),
  \label{eq:branches}
\end{align}
where $\mathrm{SE}$ is squeeze--excitation channel attention, $\mathrm{DW}_k$ a
depthwise convolution and $\mathrm{PW}$ a pointwise convolution. The
low band receives channel attention and large support to address the
large-scale, channel-structured colour and illumination variations that
predominantly appear there. The detail-bearing high band receives a small
kernel and no channel-global reweighting, which could suppress already weakened
fine structure. A gated
recombination~\cite{cui2023sfnet} weighs the bands per channel,
\begin{equation}
  \begin{aligned}
    \mathbf{v} & = \mathrm{PW}_{\mathrm{out}}\!\left(
      \alpha_l \odot \mathbf{u}_l + \alpha_h \odot \mathbf{u}_h\right), \\
    [\alpha_l,\alpha_h] & = \sigma\!\big(\mathrm{FC}(\mathrm{GAP}
      (\mathbf{u}_l{+}\mathbf{u}_h))\big),
  \end{aligned}
  \label{eq:fuse}
\end{equation}
with $\sigma$ a softmax over the concatenated per-channel attention logits of
the two branches, and the module closes residually through a
GDFN:
\begin{equation}
  \mathrm{S}(\mathbf{z}) \;=\;
  \mathrm{G}_{\mathrm{post}}\!\left(\mathbf{v} + \mathbf{z}\right).
  \label{eq:spark}
\end{equation}

\subsection{\texorpdfstring{\raggedright}{}\past{}: removable nonlinear frequency
\texorpdfstring{\mbox{scaffolding}}{scaffolding}}
\label{sec:method_past}

Building on the reserved post-attention slot, \past{} turns \spark{} into
removable nonlinear training scaffolding. Equations
\eqref{eq:split}--\eqref{eq:fuse} are nonlinear, so \spark{} admits no absorbing
algebra and cannot be collapsed into neighbouring layers as a linear branch
can~\cite{ding2021repvgg}. \past{} makes removability architectural instead.
Because the entire apparatus modulates a \emph{residual} around the slot input
(Eq.~\eqref{eq:spark}), deleting it collapses $\mathrm{S}(\mathbf{z})$ onto
$\mathrm{G}_{\mathrm{post}}(\mathbf{z})$---the permanent scaffold-free
block---up to the constant bias $\mathbf{b}_{\mathrm{out}}$ of
$\mathrm{PW}_{\mathrm{out}}$, which folds into the affine bias of the preceding
LayerNorm:
\begin{equation}
  \mathbf{b}_{\mathrm{LN}}' \;=\; \mathbf{b}_{\mathrm{LN}} +
  \mathbf{b}_{\mathrm{out}}.
  \label{eq:fold}
\end{equation}
\looseness=-1
The bias fold is algebraically exact and numerically indistinguishable in
floating-point execution (Supplementary). The inference network is then,
layer for layer and parameter for parameter, \netname{}'s permanent graph:
$28{,}586{,}687$ training parameters become $24{,}088{,}479$, and $146.86$
GFLOPs at $256^2$ become $132.15$. \past{} therefore trains the permanent
\netname{} graph under richer band-specific capacity while leaving none of the
in-block frequency machinery at inference. It is the training mechanism that
makes \netname{}'s nonlinear band-repair scaffold removable.

\subsection{Low-frequency guidance into the encoder}
\label{sec:method_lf}

\looseness=-1
To deliver the global colour cast directly to encoder stages that represent
global structure, \netname{} injects a learnable low-pass view of the input at
deeper resolutions. The encoder levels that carry global structure meet the
input only through cascaded downsampling, by which point the cast is entangled
with content. A $31\times31$ depthwise Gaussian, initialised at $\sigma=10$ and
trained end to end, extracts $\mathbf{I}_{\mathrm{lf}}$; downsampled copies at
$1/2$ and $1/4$ are encoded by shallow convolutional modules and merged into
encoder levels 2 and 3 by feature-attention
fusion~\cite{cho2021mimo,cui2023sfnet}. The path therefore supplies the
low-band problem where the network can act on it globally, complementing the
block-level design at the encoder level.

\subsection{Three-scale spatial-frequency objective}
\label{sec:method_loss}

To align supervision with the same band-asymmetric prior, \netname{}
optimises three decoder outputs in both the spatial and the frequency domain.
Decoder levels 3, 2 and 1 each carry an output head producing a residual
prediction at $1/4$, $1/2$ and full resolution. With $\hat{\mathbf{y}}_s$ the
prediction and $\mathbf{y}_s$ the reference at scale $s$,
\begin{equation}
  \mathcal{L} \;=\; \sum_{s}
  \big\|\hat{\mathbf{y}}_s-\mathbf{y}_s\big\|_1
  \;+\; \lambda \sum_{s}
  \big\|\mathcal{T}(\hat{\mathbf{y}}_s)-\mathcal{T}(\mathbf{y}_s)\big\|_1,
  \label{eq:loss}
\end{equation}
\looseness=-1
where $\mathcal{T}$ stacks the real and imaginary parts of the 2D FFT and
$\lambda=0.1$, following the multi-scale spatial-plus-frequency supervision
common in restoration~\cite{cho2021mimo,cui2023sfnet}. The objective therefore
supervises global appearance and local structure across the complete decoder
hierarchy and completes \netname{}'s architecture-wide coordination at the
objective level.

\begin{figure*}[ht]
  \centering
  \includegraphics[width=0.90\linewidth]{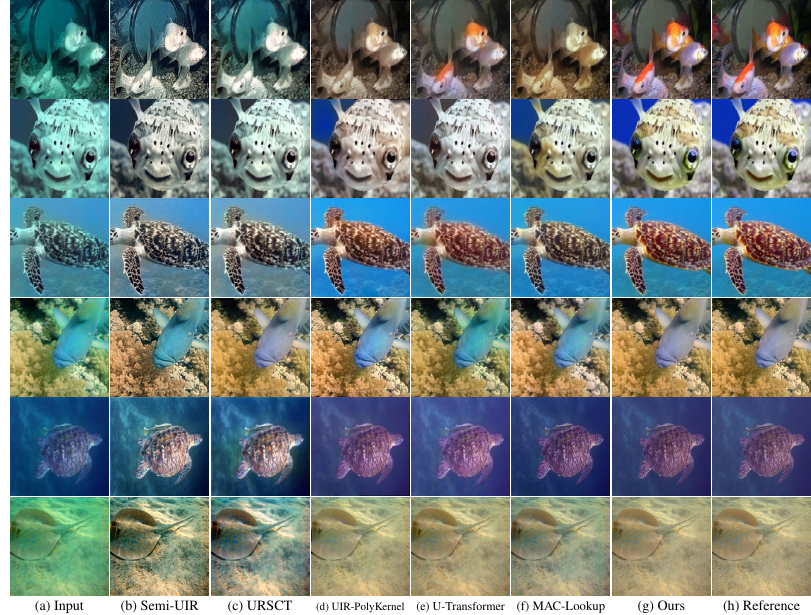}
  \caption{Samples for visual comparison showcase how SandBAR achieves more
  faithful colour restoration and clearer structural detail by repairing the
  low- and high-frequency bands with separate operators, while the compared
  baselines either retain the input's cast or soften surviving detail.}
  \label{fig:qualitative}
\end{figure*}
\section{Experiments}
\label{sec:experiments}

\subsection{Data and evaluation}
\label{sec:exp_details}

We follow the public 4{,}800-pair underwater restoration
benchmark~\cite{polykernel2025} (800 UIEB~\cite{li2019underwater} + 2{,}000
EUVP~\cite{islam2020fast} + 2{,}000 LSUI~\cite{peng2023ushape}) and evaluate on
its corresponding 490-image paired test set, comprising 200 EUVP, 200 LSUI and
90 UIEB images. \netname{} uses an encoder--decoder with dim $32$, blocks
$(2,4,4,6)$, heads $[1,2,4,8]$ and expansion $2.66$. Final models are trained
for $650$ epochs with Adam on random $256\times256$ crops using a batch size of
$4$. The learning rate is warmed up for three epochs and then cosine-annealed
from $10^{-4}$ to $10^{-6}$. Controlled architectural ablations in
Tables~\ref{tab:placement} and~\ref{tab:ablation} use the same training
configuration with a $100$-epoch budget. Complete training settings are
provided in the supplementary material. Cost figures are measured on one RTX
5090; the complete timing procedure is also provided in the supplementary
material.

\begin{table*}[t]
  \centering
  \caption{\textbf{Comparison on the 4{,}800-pair underwater restoration
  benchmark} (published results from the source
  papers~\cite{deepsea2025,maclookup2025}).}
  \label{tab:family}
  \footnotesize
  \setlength{\tabcolsep}{4pt}
  \begin{tabular*}{\textwidth}{@{\extracolsep{\fill}}l cc cc c c@{}}
    \toprule
    & \multicolumn{2}{c}{EUVP} & \multicolumn{2}{c}{LSUI} & UIEB & 3-domain \\
    \cmidrule(lr){2-3}\cmidrule(lr){4-5}\cmidrule(lr){6-6}
    Method & PSNR$\uparrow$ & SSIM$\uparrow$ & PSNR$\uparrow$ & SSIM$\uparrow$ &
      PSNR$\uparrow$ & mean PSNR$\uparrow$ \\
    \midrule
    ROP~\cite{rop2022} & 15.34 & 0.714 & 17.38 & 0.806 & 18.48 & 17.07 \\
    ADPCC~\cite{adpcc2023} & 15.20 & 0.692 & 16.20 & 0.763 & -- & -- \\
    WWPF~\cite{wwpf2023} & 15.95 & 0.648 & 17.90 & 0.739 & -- & -- \\
    UIE-WD~\cite{uiewd2022} & 17.80 & 0.760 & 19.23 & 0.803 & -- & -- \\
    TACL~\cite{tacl2022} & 20.99 & 0.782 & 22.97 & 0.828 & -- & -- \\
    USUIR~\cite{usuir2022} & 21.94 & 0.810 & 23.75 & 0.860 & -- & -- \\
    PUGAN~\cite{pugan2023} & 22.58 & 0.820 & 23.14 & 0.836 & 20.52 & 22.08 \\
    SFGNet~\cite{sfgnet2024} & 22.68 & 0.585 & 22.71 & 0.653 & 19.57 & 21.65 \\
    TUDA~\cite{tuda2023} & 23.73 & 0.843 & 25.52 & 0.878 & 22.72 & 23.99 \\
    Semi-UIR~\cite{huang2023semiuir} & 24.59 & 0.821 & 25.40 & 0.843 & 23.64 & 24.54 \\
    GUPDM~\cite{gupdm2023} & 24.79 & 0.847 & 25.33 & 0.877 & -- & -- \\
    CLUIE-Net~\cite{cluienet2022} & 24.85 & 0.844 & 23.57 & 0.864 & -- & -- \\
    U-Transformer~\cite{peng2023ushape} & 24.99 & 0.829 & 25.15 & 0.838 & 20.75 & 23.63 \\
    FUnIE-GAN~\cite{islam2020fast} & 22.62 & 0.720 & 21.66 & 0.744 & 19.17 & 21.15 \\
    URSCT~\cite{ursct2022} & 25.74 & 0.855 & 25.87 & 0.883 & -- & -- \\
    Ucolor~\cite{li2021underwater} & 23.72 & -- & 21.30 & -- & 18.37 & 21.13 \\
    UIR-PolyKernel~\cite{polykernel2025} & 26.42 & 0.866 & 26.55 & 0.888 & 23.52 & 25.50 \\
    MAC-Lookup~\cite{maclookup2025} & 26.85 & 0.891 & 27.78 & 0.931 & 23.84 & 26.16 \\
    DEEP-SEA~\cite{deepsea2025} & 29.27 & 0.905 & 28.90 & 0.897 & -- & -- \\
    \midrule
    \netname{} w/o \past{} & \PlainEUVPPSNR{} & \PlainEUVPSSIM{} &
      \PlainLSUIPSNR{} & \PlainLSUISSIM{} & \PlainUIEBPSNR{} & \PlainMeanPSNR{} \\
    \textbf{\netname{} (ours)} & \textbf{\OursEUVPPSNR{}} &
      \textbf{\OursEUVPSSIM{}} & \textbf{\OursLSUIPSNR{}} & \textbf{\OursLSUISSIM{}} &
      \OursUIEBPSNR{} & \textbf{\OursMeanPSNR{}} \\
    \bottomrule
  \end{tabular*}
\end{table*}

\subsection{State-of-the-art restoration}
\label{sec:exp_main}

As a complete system, \netname{} establishes the strongest three-domain mean
PSNR on the benchmark
family, reaching \OursMeanPSNR{}\,dB and exceeding the strongest published
result by approximately \PublishedMeanMargin{}\,dB (Table~\ref{tab:family}). It
reaches \OursEUVPPSNR{}\,dB on EUVP and \OursLSUIPSNR{}\,dB on LSUI.
MAC-Lookup~\cite{maclookup2025} obtains a higher LSUI SSIM of $0.931$ versus
our \OursLSUISSIM{}, using explicit SSIM supervision together with stacked
multi-stage local/global refinement. The resulting system is comparatively
compute-heavy at $173.2$\,GMACs, and this structural-similarity advantage is
accompanied by a clear reconstruction-fidelity trade-off: its LSUI PSNR is
$27.78$\,dB, $1.14$\,dB below \netname{}'s \OursLSUIPSNR{}\,dB. \netname{}
therefore delivers substantially higher pixel-level fidelity while retaining
competitive structural similarity.

\subsection{Zero-cost inference through \past{}}
\label{sec:exp_past}

We next isolate \past{}, the removable-scaffolding contribution within
\netname{}, from the complete architecture.

\begin{table}[t]
  \centering
  \caption{\textbf{\past{} inference accounting.} The \spark{}-retained row
  evaluates the trained \netname{} model before strip-and-fold, with the
  frequency scaffold kept at inference. After strip-and-fold, \netname{}
  recovers the permanent graph's parameter and FLOP counts and matched latency,
  while preserving its three-domain mean PSNR.}
  \label{tab:deploy}
  \footnotesize
  \setlength{\tabcolsep}{1.5pt}
  \begin{tabular*}{\linewidth}{@{\extracolsep{\fill}}l r r r r r@{}}
    \toprule
    & Params$\downarrow$ & GFLOPs$\downarrow$ &
      \multicolumn{2}{c}{Latency (ms)$\downarrow$} & 3-domain \\
    \cmidrule(lr){4-5}
    Graph & & @256 & @256 & @512 & mean PSNR$\uparrow$ \\
    \midrule
    \spark{} retained & 28{,}586{,}687 & 146.86 & 27.62 & 81.52 & 27.20 \\
    \textbf{Post-fold} & 24{,}088{,}479 &
      132.15 & 21.53 & 61.89 & 27.19 \\
    \netname{} w/o \past{} & 24{,}088{,}479 & 132.15 & 21.44 & 61.91 & 26.79 \\
    \bottomrule
  \end{tabular*}

  \vspace{2pt}
  {\raggedright\footnotesize Peak memory$\downarrow$: $489.0$ / $488.2$\,MB at $256^2$ for
  the strip-and-fold \past{} graph and the scaffold-free baseline. Fold precision in
  Supplementary.\par}
\end{table}

\looseness=-1
\past{} places the complete cost of \spark{} in training. Removing the scaffold
reduces the graph from $28{,}586{,}687$ to $24{,}088{,}479$ parameters and from
$146.86$ to $132.15$ GFLOPs, while the strip-and-fold \past{} model matches the permanent
\netname{} graph in latency at both resolutions (Table~\ref{tab:deploy}).
The first row evaluates the trained \netname{} model before strip-and-fold,
with \spark{} retained at inference; it is not the graph used for the reported
\netname{} inference results. Retaining \spark{} gives a three-domain mean PSNR
of $27.20$\,dB, whereas the post-fold inference graph gives $27.19$\,dB. The
unrounded values differ by only $0.0036$\,dB, showing that strip-and-fold
preserves restoration performance while removing the frequency scaffold from
the inference graph.

\begin{figure}[t]
  \centering
  \includegraphics[width=\linewidth,trim=0bp 0.3bp 0bp 10bp,clip]{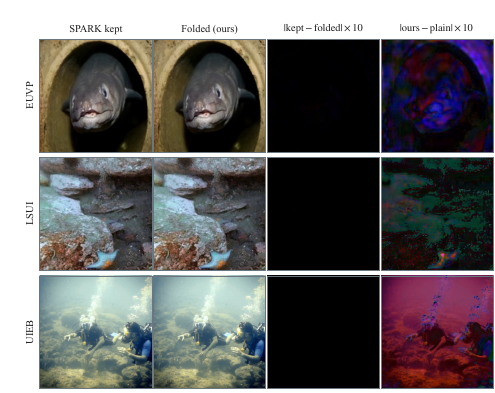}
  \caption{\textbf{Removing \spark{} at inference is visually inert; training
  with it is not.} The first two columns evaluate the trained \netname{} model
  before and after strip-and-fold. The third column shows their
  absolute difference amplified $10\times$; the fourth applies the same
  amplification to the difference between the post-fold graph and the plain
  baseline. Depending on the domain, the mean absolute difference is
  $20$--$139\times$ larger in the fourth column, so the near-black third column
  is a property of strip-and-fold rather than insufficient amplification.}
  \label{fig:invariance}
\end{figure}

\begin{figure}[t]
  \centering
  \includegraphics[width=\linewidth,trim=0bp 4.3bp 0bp 13bp,clip]{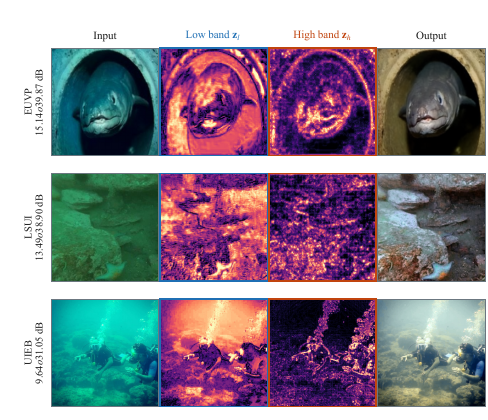}
  \caption{\textbf{What the band scaffold learns.} Channel-mean magnitudes of
  the \spark{} band split ($\mathbf{z}_l$ and $\mathbf{z}_h$ columns; each map
  normalised independently) on one test image from each domain, captured from
  the \emph{training} graph at the second encoder stage. The split is exact
  (Eq.~\eqref{eq:split}): the low branch carries smooth illumination and
  colour-cast structure, while the high branch isolates contours, bubble
  trails and rock texture. After strip-and-fold this machinery is absent at
  inference (Sec.~\ref{sec:method_past}); the plain graph that remains
  improves 472 of the 490 test images in both bands (Gaussian split,
  $\sigma{=}10$).}
  \label{fig:bandfeat}
\end{figure}

Figure~\ref{fig:invariance} makes the two effects visually distinct.
Strip-and-fold changes the trained model's output negligibly, whereas training
with the scaffold changes the permanent graph substantially: the
post-fold--plain mean absolute difference is $20.1\times$, $53.8\times$ and
$139.2\times$ larger on
EUVP, LSUI and UIEB, respectively, than the retained--post-fold difference.
Despite identical inference cost, \past{} improves PSNR on all three domains:
by \PaSTEUVPGain{}, \PaSTLSUIGain{} and \PaSTUIEBGain{}\,dB on EUVP, LSUI and
UIEB, a three-domain mean of \PaSTMeanGain{}\,dB. These gains are computed from
the underlying unrounded scores of the last two rows of Table~\ref{tab:family}.
\textbf{No part of the frequency scaffold is present in the post-fold inference
graph.} These results demonstrate that removable nonlinear frequency capacity
strengthens the permanent restorer without entering its inference graph.
Figure~\ref{fig:bandfeat} visualises the split the scaffold learns before it
is removed: the low branch concentrates the colour cast, the high branch the
fading detail.

\begin{table}[t]
  \centering
  \caption{\textbf{Effect of band decomposition.} The unsplit variant retains the two repair branches. Scores are equal-domain means over EUVP, LSUI and UIEB, evaluated from floating-point predictions after scaffold removal.}
  \label{tab:band_decomposition}
  \footnotesize
  \begin{tabular*}{\linewidth}{@{\extracolsep{\fill}}lcc@{}}
    \toprule
    Training scaffold & PSNR$\uparrow$ & SSIM$\uparrow$ \\
    \midrule
    Without band decomposition & 27.069 & 0.91248 \\
    \textbf{SandBAR} & \textbf{27.193} & \textbf{0.91493} \\
    \bottomrule
  \end{tabular*}
\end{table}

With the same permanent inference architecture, SandBAR exceeds the unsplit variant by 0.124 dB PSNR and 0.00245 SSIM.

\subsection{Where band repair belongs}
\label{sec:exp_slot}

The preceding experiment establishes that band-specific capacity can shape a
stronger permanent graph without remaining at inference. We next ask why that
capacity is assigned specifically to the post-attention slot. Table~\ref{tab:placement}
isolates placement from module and backbone identity. With \spark{} in
\netname{}, moving band repair from the pre-attention slot to the
post-attention slot raises the three-domain mean from $26.43$ to
$26.72$\,dB. Replacing \spark{} with DFFN~\cite{kong2023fftformer}
reproduces the effect, from $26.25$ to $26.52$\,dB. Changing the backbone
instead gives the same result: \spark{} in Restormer~\cite{zamir2022restormer}
improves from $26.15$ to $26.44$\,dB when moved after attention. Across all
three controlled comparisons, post-attention placement contributes
$0.27$ to $0.29$\,dB. The gain therefore follows the slot across two
different modules and two different backbones.

\begin{table}[t]
  \centering
  \caption{\textbf{Where band repair belongs.}
  Three-domain mean PSNR under the controlled $100$-epoch budget.
  Each row compares the same module in the pre-attention and post-attention
  feed-forward slots of one backbone, with all other settings held fixed.
  \spark{} and \netname{} are ours, while DFFN and Restormer are existing
  designs.}
  \label{tab:placement}
  \footnotesize
  \begin{tabular*}{\linewidth}{@{\extracolsep{\fill}}llcc@{}}
    \toprule
    \multicolumn{2}{c}{\textbf{Configuration}} &
      \multicolumn{2}{c}{\textbf{Mean PSNR$\uparrow$}} \\
    \textbf{Module} & \textbf{Backbone} &
      \textbf{pre-attn.} & \textbf{post-attn.} \\
    \midrule
    \spark{} & \netname{} &
      26.43 & \textbf{26.72} \\
    DFFN~\cite{kong2023fftformer} & \netname{} &
      26.25 & \textbf{26.52} \\
    \spark{} & Restormer~\cite{zamir2022restormer} &
      26.15 & \textbf{26.44} \\
    \bottomrule
  \end{tabular*}
\end{table}

\looseness=-1
The result exposes a functional asymmetry between the two feed-forward
positions. The pre-attention FFN operates before global context has been
assembled, and any frequency-specific transformation it produces is
subsequently re-mixed by attention. The post-attention FFN receives the
context-aware representation instead and can translate the global view
directly into restoration. For underwater degradation, this ordering is
particularly important: scene-level evidence establishes the global colour,
haze and illumination state first, after which band-specific processing can
apply large-support, channel-aware correction to the low band and local
detail-preserving refinement to the high band. \netname{} therefore assigns
generic mixing to the front FFN and reserves the rear FFN for band repair.
The sandwich block is not simply a block with an additional FFN; its ordering
implements the architectural principle of looking globally first and repairing
by frequency afterwards.

For Split and the unsplit variant, post-attention placement changes mean PSNR by $+0.290$ and $+0.139$ dB with global attention, and $-0.011$ and $-0.158$ dB with windowed attention.

\subsection{Architectural ablation}
\label{sec:exp_ablation}

\begin{table}[t]
  \centering
  \caption{\textbf{Architectural ablation.}
  Model (a) is the full \netname{} scaffold. All variants are trained under the
  same controlled $100$-epoch budget as Table~\ref{tab:placement}, with each
  variant changing exactly one design decision and all other settings held fixed.
  $\Delta$ is the three-domain mean PSNR relative to model (a).}
  \label{tab:ablation}
  \footnotesize
  \setlength{\tabcolsep}{1.25pt}
  \begin{tabular*}{\linewidth}{@{\extracolsep{\fill}}cl ccc r r@{}}
    \toprule
    & Variant & Splitter & Band ops & LF guid. & PSNR$\uparrow$ &
      $\Delta\uparrow$ \\
    \midrule
    (a) & \netname{} full        & adaptive & asym. & \checkmark &
      \textbf{\AblFullPSNR{}} & --- \\
    (b) & fixed Gaussian         & fixed    & asym. & \checkmark &
      \AblGsplitPSNR{} & $-\AblGsplitDropPct{}\%$ \\
    (c) & symmetric ops          & adaptive & sym.  & \checkmark &
      \AblSymopsPSNR{} & $-\AblSymopsDropPct{}\%$ \\
    (d) & no LF guidance         & adaptive & asym. & --- &
      \AblNoLFPSNR{} & $-\AblNoLFDropPct{}\%$ \\
    \bottomrule
  \end{tabular*}
\end{table}

We next isolate the remaining constituent designs. Under the controlled
$100$-epoch setting of Tables~\ref{tab:placement} and~\ref{tab:ablation},
model~(a) is the full \netname{} scaffold. Replacing
its content-adaptive splitter with a fixed Gaussian split in model~(b) degrades
PSNR by
$\AblGsplitDropPct{}\%\downarrow$ and SSIM by
$\AblGsplitSSIMDropPct{}\%\downarrow$, on every domain. This is because a fixed
kernel cuts at a spatial frequency the image does not respect: it leaves the
colour cast in the high band and structure in the low band, so the asymmetric
operators are applied to mismatched content.
Giving both bands the stronger low-band operator in model~(c) costs
$\AblSymopsDropPct{}\%\downarrow$ in PSNR, demonstrating the benefit of
matching each operator to the dominant content of its band rather than
enlarging both. Removing the low-frequency guidance path in model~(d) costs
$\AblNoLFDropPct{}\%\downarrow$---as much as the asymmetry itself---while the
same path is worth $\AblGuidanceAlone{}$\,dB on the permanent graph alone,
indicating that guidance pays off only where band-specific repair can act on
what it delivers. These results confirm that \netname{}'s three levels operate
as one system across diverse underwater scenarios rather than as independent
additions.

\looseness=-1
In the architectural ablations, floating-point evaluation gives 26.721 dB for the archived full model, 26.168 dB without FFT loss, and 26.477 dB with full-resolution supervision alone.
The interaction between SPARK and LF guidance is $+0.242$ dB in mean PSNR without FFT supervision.

\subsection{Qualitative results}
\label{sec:exp_qual}

Beyond the quantitative gains, SandBAR shows a clear visual advantage over the
compared methods. As shown in Figure~\ref{fig:qualitative}, separating each
feature map into a low- and a high-frequency band and repairing the two with
different operators lets the model correct the global colour shift without
flattening the detail that survives the water column. The compared methods
recover scene structure but not its colour: some carry the input's cast
through to the output, others desaturate exactly the regions the reference
renders most vividly. Our results keep the reference's colour distribution,
preserve object boundaries and retain fine texture, which together indicate
that band-asymmetric repair is what the visual improvement rests on.

\section{Conclusion}
\label{sec:conclusion}

\looseness=-1
Our main contribution is \netname{}, a novel unified band-asymmetric
architecture for underwater image restoration. It turns the band-asymmetric
tendency of large-scale colour distortion and local high-frequency detail loss
into coordinated architectural targets across block topology, encoder guidance
and three-scale spatial-frequency supervision. Within this architecture, the
novel band-asymmetric sandwich block reserves a post-attention repair slot, and
the novel \spark{} module applies channel-aware large-support correction to the low band and
lightweight local refinement to the high band, restoring global colour while
preserving fine structure. To make this nonlinear band-repair capacity
removable, \past{} residual-anchors the scaffold and folds its remaining bias.
The strengthened permanent \netname{} graph therefore runs at inference without
the frequency scaffold, with unchanged parameters and FLOPs at matched latency.
\netname{} establishes a new
three-domain mean state of the art on the benchmark
\mbox{family}, exceeding the strongest published result by approximately
\PublishedMeanMargin{}\,dB. These results establish band asymmetry as an
effective architectural prior for underwater restoration and removable
nonlinear scaffolding as a promising route to stronger inference-efficient
models. Extending removable
nonlinear scaffolding across restoration tasks is an important direction for
future work.

\clearpage
\clearpage
{
  \small
  \bibliographystyle{ieeenat_fullname}
  \bibliography{refs}
}

\end{document}